\documentclass[sn-basic,iicol]{sn-jnl}
\usepackage[latin1]{inputenc}
\usepackage{stmaryrd}
\usepackage{lineno,hyperref}
\usepackage{bbm}
\usepackage{bm}
\usepackage{amsfonts}
\usepackage{amsmath}
\usepackage{mathrsfs}
\usepackage{amssymb}
\usepackage{enumerate}
\usepackage{graphicx}
\usepackage{subfigure}
\usepackage{booktabs}
\usepackage{graphicx}
\usepackage{setspace}
\usepackage[nodots]{numcompress}
\usepackage{algorithm}
\usepackage{algorithmicx}
\usepackage{algpseudocode}
\usepackage{setspace}
\usepackage{multirow}
\usepackage{color}
\usepackage{float}
\usepackage{array}
\usepackage{geometry}
\usepackage{times}
\usepackage{mathptmx}
\usepackage{dsfont}
\usepackage{soul}
\usepackage{url}

\jyear{2021}%

\theoremstyle{thmstyleone}%
\theoremstyle{thmstyletwo}%

\theoremstyle{thmstylethree}%

\begin{document}

\title[Article Title]{Adversarial Self-Attack Defense and Spatial-Temporal Relation Mining for Visible-Infrared Video Person Re-Identification}


\author[1]{\fnm{Huafeng} \sur{Li}}\email{hfchina99@163.com}
\equalcont{These authors contributed equally to this work.}

\author[1]{\fnm{Le} \sur{Xu}}\email{xuleleo2022@163.com}
\equalcont{These authors contributed equally to this work.}

\author*[1]{\fnm{Yafei} \sur{Zhang}}\email{zyfeimail@163.com}

\author[2]{\fnm{Dapeng} \sur{Tao}}\email{dapeng.tao@gmail.com}

\author[1]{\fnm{Zhengtao} \sur{Yu}}\email{ztyu@hotmail.com}

\affil[1]{\orgdiv{Faculty of Information Engineering and Automation}, \orgname{Kunming University of Science and Technology}, \orgaddress{\city{Kunming}, \postcode{650500}, \state{Yunnan}, \country{China}}}
\affil[2]{\orgdiv{School of Information Science and Engineering}, \orgname{Yunnan University}, \orgaddress{\city{Kunming}, \postcode{650091}, \state{Yunnan}, \country{China}}}

\abstract{In visible-infrared video person re-identification (re-ID), extracting features not affected by complex scenes (such as modality, camera views, pedestrian pose, background, etc.) changes, and mining and utilizing motion information are the keys to solving cross-modal pedestrian identity matching. To this end, the paper proposes a new visible-infrared video person re-ID method from a novel perspective, i.e., adversarial self-attack defense and spatial-temporal relation mining. In this work, the changes of views, posture, background and modal discrepancy are considered as the main factors that cause the perturbations of person identity features. Such interference information contained in the training samples is used as an adversarial perturbation. It performs adversarial attacks on the re-ID model during the training to make the model more robust to these unfavorable factors. The attack from the adversarial perturbation is introduced by activating the interference information contained in the input samples without generating adversarial samples, and it can be thus called adversarial self-attack. This design allows adversarial attack and defense to be integrated into one framework. This paper further proposes a spatial-temporal information-guided feature representation network to use the information in video sequences. The network cannot only extract the information contained in the video-frame sequences but also use the relation of the local information in space to guide the network to extract more robust features. The proposed method exhibits compelling performance on large-scale cross-modality video datasets. The source code of the proposed method will be released at \textcolor{blue}{\url{https://github.com/lhf12278/xxx}}.}




\maketitle
\section{Introduction}\label{sec1}
Person re-identification (re-ID) is a technology used to determine whether the person images or sequences captured by non-overlapping cameras belong to the same identity~\cite{1, 2, 3, 4, 69, 71, 72, 73, 78, 79, 80}. Most of the related works, such as domain generalization~\cite{5,6,7,8} and domain adaptation\cite{9,10,11,12,62,63,64}, focus on person re-ID under normal illumination (visible modality). In recent years, cross-modality person re-ID~\cite{13,14,15,16,17,70,75,76,77} based on visible and infrared images have attracted more and more attention since they can meet the requirements of pedestrian image matching under poor illumination at night.
\begin{figure*}[t!]
	\centering
	\includegraphics[width=1.0\textwidth]{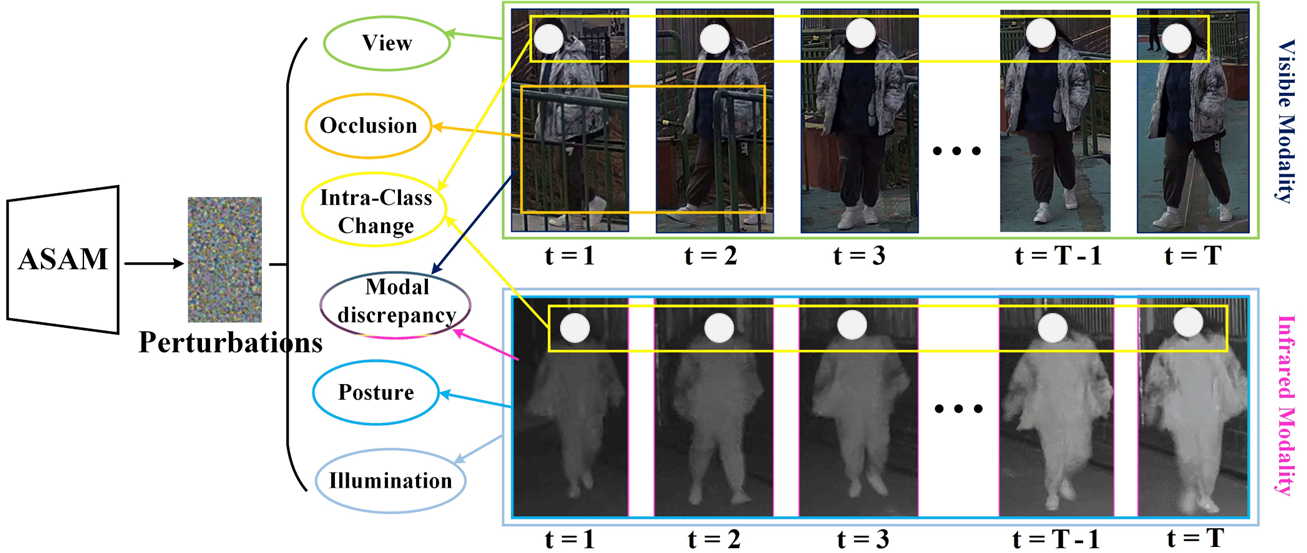}
	\caption{Illustration of perturbations existing in cross-modality person images. All factors that lead to changes in pedestrian identity features are treated as adversarial perturbations. The pedestrian identity matching performance would be improved by enhancing the model's defense against these perturbations.}
	\label{fig1}
\end{figure*}

The main difficulty faced by visible-infrared person re-ID is the modality discrepancy between pedestrian images of two modalities. Most methods~\cite{18,19,15,21,22,23,24,25,66} attempt to study how the features can be learned without being affected by the modality of images. These methods can be roughly divided into methods based on adversarial learning~\cite{18,19,15}, methods guided by intermediate modality~\cite{21,22,23}, and methods embedded with high-level semantic information~\cite{24,25,66}. The adversarial learning-based methods achieve modal confusion by adversarial training between the encoder and the discriminator, thereby reducing the difference between different modalities. The intermediate modality-based methods use the information of intermediate modality to guide or strengthen the role of modality invariant features in identity matching. The semantic information-based methods improve the cross-modality capability of features by introducing high-level semantic information into visual features.

The above methods only consider modal discrepancy's impact on person identity matching. However, some aspects are ignored, such as the diversity of pedestrian appearance features caused by view discrepancies, diverse postures of person, and background changes, etc. Moreover, they are designed for the matching between person images and do not consider the information contained in the video sequences. Therefore, if the existing cross-modality person re-ID methods are directly applied to the cross-modality video person re-ID, the retrieval performance may not be optimal. Although video-based person re-ID methods~\cite{26,27,28,29} are widely applied, they usually do not consider the relationship between different parts of the pedestrian's body under the motion state, which limits the further improvement of the recognition performance. Furthermore, most of the existing video person re-ID methods focus on the identity matching between video sequences under normal lighting conditions, ignoring the impact of the modal discrepancy between infrared and visible person images. Although Lin et al.~\cite{30} proposed a video-based cross-modality person re-ID method and created the first dataset for this task recently, there are still few studies involving the solution to this problem.

We propose a cross-modality video person re-ID method from a novel view---adversarial self-attack and defense. In the proposed method, we regard all unfavorable factors contaminating the model performance as adversarial perturbations. Factors such as the change of camera view, the existence of occluders, the difference in posture, and the gap between modalities lead to a certain diversity of appearance features of the same identity, as illustrated in Fig.~\ref{fig1}.

We regard all the differences in appearance features of the same identity caused by all factors as information perturbations. The robustness of the re-ID model is strengthened by improving its defense ability against perturbations. Technically, the proposed method is mainly composed of the adversarial self-attack module (ASAM),  adversarial defense module (ADM) and feature representation module under spatial-temporal information guidance (FRM-STIG). The  ASAM is mainly used to activate the adversarial perturbations and implement the attack on the ADM. More specifically, the ASAM is used to guide the single-modality feature extraction network to activate the perturbations in the input training samples. With the effect of ASAM,  the robustness of the ADM is enhanced in adversarial training of the ADM. The proposed method does not need to synthesize adversarial samples to train the model but activates the adversarial perturbations of the training samples to realize the adversarial attack.

In FRM-STIG, considering the discrimination of the spatial relationship of different body parts under the motion state, we propose to embed the temporal information contained in the video frames into spatial relations of different body parts. To effectively utilize spatial relation to improve the discrimination of features, we propose a spatial-temporal relation-guided feature representation method. More attention is paid to the features related to motion information and spatial relation. Thanks to this design, both the spatial relation of different body parts during motion and the temporal information can be embedded into the pedestrian features, which helps to improve the accuracy and robustness of person video sequence description. Finally, the features with motion information and the features generated by the guidance of spatial-temporal relations are combined as the final features to describe pedestrians.

The main contributions of this paper are as follows:

\begin{itemize}
	\item A solution is proposed to solve the impact of modal discrepancy, posture changes, complex background and other factors on person identity matching. The complex perturbations carried by the multi-modality images are treated as the adversarial attack information of the re-ID model. At the same time, by improving the defense ability of the re-ID model against these perturbations, the robustness of the model to complex factors can be improved accordingly.
	
	\item An adversarial self-attack strategy is proposed to activate the perturbation information contained in the input samples without generating adversarial samples. This design allows adversarial attack and defense to be integrated into one framework.
	
	\item A spatial relation mining mechanism is proposed for different parts of a person based on temporal information embedding. A feature highlight mechanism guided by spatial-temporal relations is designed to construct features not affected by modality.
	
	\item The validity of the proposed method is verified on the challenging large-scale visible-infrared video re-ID dataset---VCM and the state-of-the-art performance is obtained under two commonly used evaluation metrics.
\end{itemize}

The rest of this paper is organized as follows. Section~\ref{sec:related} discusses the related state-of-the-art works. Section~\ref{sec:method} elaborates the proposed method in detail. The experimental results are explained in Section~\ref{sec:experiment} and Section~\ref{sec:conclusion} summarizes the content of this paper and draws conclusions.

\section{Related Work}\label{sec2}
\subsection{Visible-Infrared Person Re-ID}
To solve the modal discrepancy between visible and infrared person images, Wang et al.~\cite{15} proposed an adversarial generation method to learn the modality-invariant feature. The encoder can extract the features not affected by the modality information via playing a min-max game between the encoder and the modality discriminator. Given the significance of adversarial learning, a series of practical methods have been conducted~\cite{31,32,33}. However, in those methods, the discriminator is used to identify the modal differences between visible and infrared person images, which may cause the loss of information related to personal identity and is not conducive to matching person identity. Another popular way to learn modality-invariant features is to use the intermediate information between two modalities as guidance~\cite{21,22,23}. Specifically, Zhong et al.~\cite{23} proposed the gray-scale image of a person as an intermediate modality to assist in extracting modality-invariant features.

Considering the modality invariance of edge details of a person image, Gao et al.~\cite{25} enhanced the cross-modal matching ability of features by highlighting the role of edge details in features. Basaran et al.~\cite{34} proposed to extract modality-invariant identity features by introducing the anaglyph. However, it ignores the high-level semantic information between different body parts or critical points of a person. Such information is usually modal invariant and often used in cross-modality person re-ID. Miao et al.~\cite{35} proposed a cross-modality person re-ID method based on high-order relationship mining of person key points. Chen et al.\cite{24} proposed a modality-invariant feature extraction method by mining different part relationships. Those methods are devoted to extracting features not affected by modality, where the challenges to person identity matching caused by the diversity of person appearance features are not considered. The proposed method is based on adversarial self-attack and defense such that the changes in personal appearance features caused by all factors are deemed adversarial perturbations. The shortcomings of the above methods can be alleviated by improving the model's robustness against such perturbations.

\subsection{Video Person Re-ID}
Videos usually contain many motion information, which carries pedestrian identity clues. The video person re-ID has received more and more attention~\cite{67,68}. Wu et al.~\cite{36} proposed 3D ConvNet as a new layer of feature extraction network to extract a person's appearance and motion information from the video sequences simultaneously. Chen et al.~\cite{37} proposed spatial-temporal awareness to pay attention to the significant parts of a person in both temporal and spatial domains simultaneously and highlight the effect of this part in identity matching. Li et al.\cite{38} proposed a global-local temporal representation (GLTR) method for video person re-ID. This method aggregates the short-term temporal cues and long-term relations as the final GLTR. Liu et al.~\cite{39} proposed a co-saliency spatial-temporal interaction network (CSTNet) for video person re-ID. The method learned discrimination feature representation by capturing the salient foreground regions in the video and exploring the spatial-temporal long-range context interdependency from such regions. Yang et al.~\cite{40} designed a two-stream dynamic pyramid representation model to solve the problems of mining spatial-temporal information, suppressing redundant information and improving data quality for video person re-ID. The method used dynamic pyramid deflated convolution and pyramid attention pooling to acquire the person's motion information and static appearance. Eom et al.~\cite{41} designed a spatial and temporal memory network to address the challenge of person occlusion by using prior knowledge that spatial distractors always appear in a particular location. In contrast, temporal distractors usually appear in the first few frames. Liu et al.~\cite{42} adopted a bi-directional (forward and backward) mechanism to extract the temporal information in the video sequence.

Although the above methods effectively utilized the motion information for person re-ID, they ignore the potential structural relationship of a person's body parts in space, limiting the further improvement of feature discrimination. Yan et al.~\cite{43} proposed multi-granular hypergraphs to mine the temporal information of different granularity regions. They modeled spatial-temporal dependencies in terms of multiple granularities, which effectively improved the performance of video person re-ID. Liu et al.\cite{44} proposed a spatial-temporal correlation multi-scale topology learning framework to realize video person re-ID. The method achieved hierarchical spatial-temporal dependencies and pedestrian structure information through 3D and cross-scale graph convolution. To solve the problem that 3D convolution is easily affected by the misalignment of person features in mining temporal information, Chen et al.~\cite{45} proposed a human-oriented graph method. Although the above methods based on graph convolution can mine the spatial relationship between nodes, they cannot extract long-term spatial cues. Since the transformer is more suitable for extracting the long-term relationship of features, Zhang et al.~\cite{46} proposed a spatial-temporal transformer for video person re-ID. The method is mainly composed of a spatial transformer and a temporary transformer. The former is used to extract the spatial features of person images, and the latter is used to extract the features of a person in video sequences. Although these methods consider the static spatial structure relation between different person regions, they ignore the discrimination of different person body parts when moving. The proposed method embeds the temporal information into the spatial structure information mining, resulting in a spatial relation mining scheme for different body parts of pedestrians in the state of motion.

\subsection{Adversarial Attack and Defense in Person Re-ID}
Adversarial attacks are designed to deprive the original performance of the deep neural network by adding small-magnitude perturbations to original samples. The concept was first proposed by Szegedy et al.~\cite{47}. Wang et al.~\cite{48} developed a multi-stage network to perform back-box attack, given the importance of cross-dataset transferability in Re-ID. It pyramids the features at different levels to extract the general and transferable features for the adversarial perturbations. To explore whether the re-ID model based on CNN is vulnerable to the attack of adversarial samples, Wang et al.~\cite{49} proposed an attack method called advPattern to generate adversarial patterns on clothes. Those methods focus on generating adversarial samples to invalidate the re-ID model without considering how to defend against attacks from adversarial samples.

One of the easiest ways to improve the re-ID model's robustness to adversarial examples is to incorporate them into training. In addition, some researchers consider identifying and excluding the adversarial samples from the training dataset via the detection algorithm, which can also avoid the attack from the adversarial samples on the model. Specifically, Wang et al.~\cite{50} proposed a multi-expert adversarial attack detection method to detect adversarial attack by checking context inconsistency. To fill the gap between training samples and test samples, Bai et al.~\cite{51} developed an adversarial metric attack while presenting an early attempt to produce a metric-preserving network, thereby protecting metrics from adversarial attacks. To defend the model against the attack of the adversarial samples, although it is simple and effective to use the adversarial samples directly to train the model, it does not maximize the robustness of the model to the adversarial samples. In this paper, we elaborately devise an adversarial self-attack and defense approach that enables the model to defend against the impact of the diversity of person identity features on matching performance. Unlike the existing methods for generating adversarial samples, the proposed method replaces the role of the adversarial samples by activating the adversarial perturbations contained in the training samples. The proposed method integrates adversarial attack and defense within a single framework.

\begin{figure*}[htbp!]
	\centering
	\includegraphics[width=1.0\textwidth]{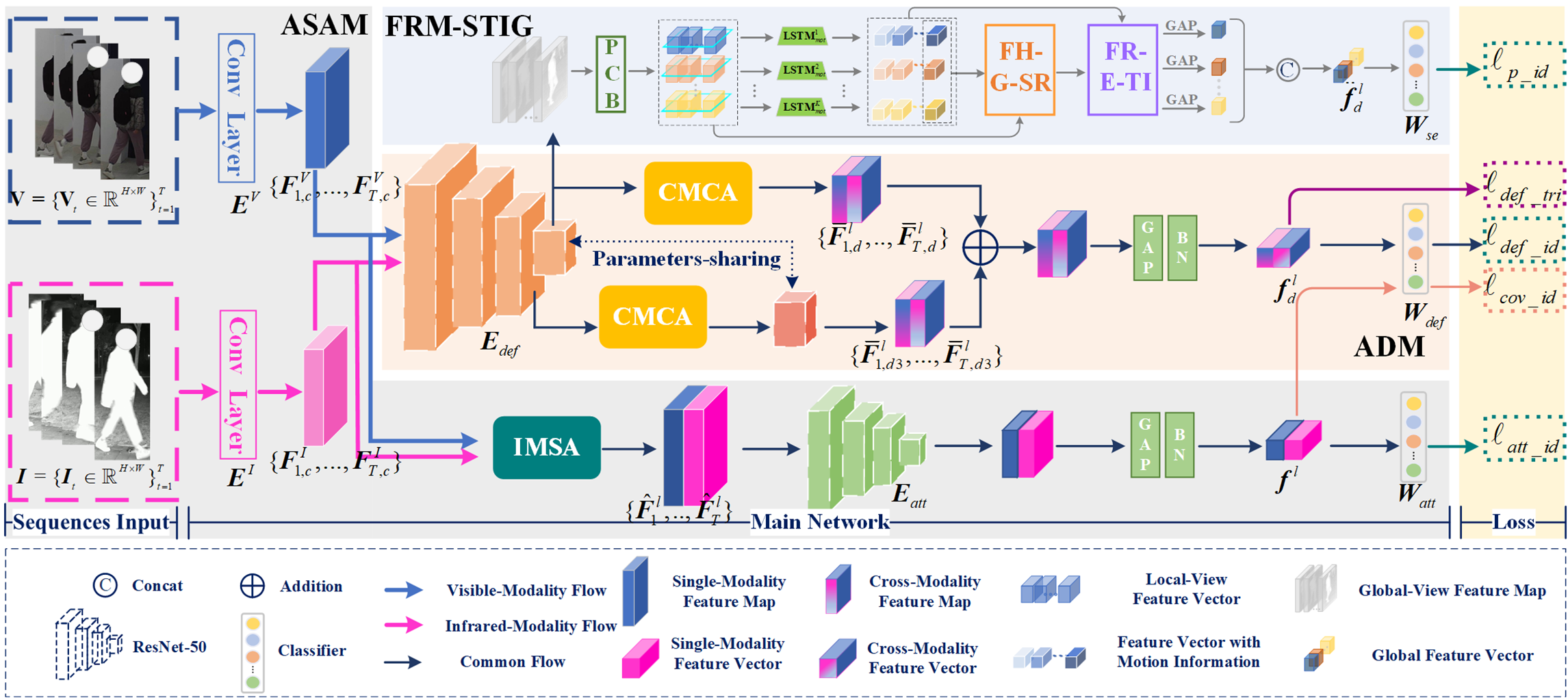}
	\caption{Overall framework of the proposed method. Visual/Infrared image sequences are sent to the adversarial self-attack module (ASAM) to activate the perturbations that will disturb the feature related to a person's identity. Then, the features carrying perturbations are fed to the adversarial defense module (ADM) to improve the defense ability of the re-ID model against perturbations. We design a feature representation module for spatial-temporal information guidance (FRM-STIG) to use the temporal information in the video sequence. The visual features are highlighted by the local spatial relation embedded with the temporal information.}
	\label{fig2}
\end{figure*}

\section{Proposed Method}\label{sec3}
\subsection{Overview}
The framework proposed in this paper is mainly composed of the adversarial self-attack module (ASAM), adversarial defense module (ADM) and feature representation module under spatial-temporal information guidance (FRM-STIG), as shown in Fig.~\ref{fig2}. The ASAM is mainly used to activate perturbations in the training samples and achieve the re-ID model's adversarial training. The ADM extracts discrimination features from the sample in which the perturbations are activated. The FRM-STIG extracts the information carried in the sequence and uses them to enhance the effect of features related to motion information. To comprehensively use the information carried in the video sequences, the FRM-STIG integrates visual features with spatial-temporal information to accurately describe a person.

\subsection{Adversarial Self-Attack Module}
The ASAM is designed to enable the training samples to replace the role of the adversarial samples. It is implemented in a single ResNet50 framework. The ASAM module contains the Conv Layer, Intra-Modality Self-Attention (IMSA) Layer, Feature Encoder $\bm E_{att}$, Global Average Pooling (GAP) Layer, and Batch Normalization (BN) Layer. The Conv Layer here refers to the first convolution layer of ResNet50. The $\bm E_{att}$ is composed of the last four layers of ResNet-50. The IMSA is used to highlight the role of the perturbations in the feature maps output by the Conv Layer. This Conv Layer generates perturbation information in the training samples to make the re-ID model yield its original performance. Compared with existing methods, ASAM does not need to generate new adversarial samples and only uses the original ones to achieve regular and adversarial training for the re-ID model. We denote the video sequence of different modality of the same identity $\bm V= \{\bm V_{t} \in \mathbb{R}^{H \times W}\} _{t = 1}^T$ and $\bm I = \{\bm I_{t}  \in \mathbb{R}^{H \times W}\} _{t = 1}^T$, where $H$ and $W$ represent the height and width of a single video frame, $\bm V$ and $\bm I$ represent a sequence of person video frames in visible and infrared modality, respectively. $T$ is the total number of frames in a sequence, $t$ means the index of the $t$-th frame. The results obtained by inputting the video sequence $\bm V$ and $\bm I$ into the Conv Layer can be expressed as:

\begin{equation}\small
	\begin{aligned}
		\bm F_{t,c}^V = \bm E^{V}(\bm V_{t}), \quad \bm F_{t,c}^I = \bm E^{I}(\bm I_t)\quad (t=1,2, \cdots, T)
	\end{aligned},
\end{equation}
where $\bm F_{t,c}^V$ and $\bm F_{t,c}^I$ denote the features output by the Conv Layer. $\bm E^{V}$ and $\bm E^{I}$ are the encoders consisting of the first convolution layer of ResNet-50, and their parameters are not shared. The encoders $\bm E^{I}$ and $\bm E^{V}$ are respectively used to extract the shallow features of visible and infrared person images.

\begin{figure}[htbp]
	\centering
	\includegraphics[width=1.0\linewidth]{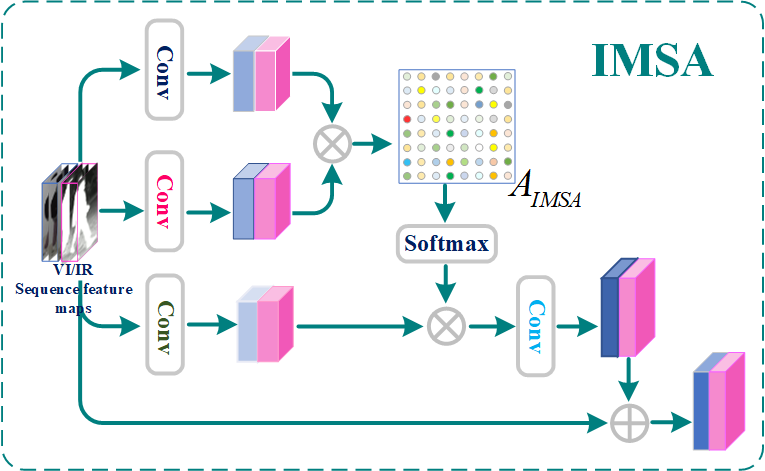}
	\caption{Structure of Intra-Modality Self-Attention (IMSA). $\otimes$ and $\oplus$ represent matrix multiplication and addition, respectively.}
	\label{fig3}
\end{figure}

To highlight the role of the perturbations carried in $\bm F_{t, c}^V$ and $\bm F_{t, c}^I$, we first send $\bm F_{t, c}^V$ and $\bm F_{t, c}^I$ to the IMSA layer, whose structure is shown in Fig.~\ref{fig3} and the output results can be expressed as:
\begin{equation}\small
	\begin{aligned}
		\hat{\bm F}_{t}^V = {\rm{IMSA}}(\bm F_{t,c}^V),\quad \hat{\bm F}_{t}^I = {\rm{IMSA}}(\bm F_{t,c}^I)
	\end{aligned}.
\end{equation}

To activate the perturbations in $\bm F_{t,c}^V$ and $\bm F_{t,c}^I$ such that they can replace the generation of adversarial examples, we send $\hat{\bm F}_{t}^V$ and $\hat{\bm F}_{t}^I$ to $\bm E_{att}$. After the perturbation information is activated, the feature of the output by the $\bm E_{att}$ followed by GAP and BN should be misclassified by the pre-trained person identity classifier $\bm W_{def}$ of ADM.
To this end, we use the following identify loss function to optimize $\bm E^{V}$, $\bm E^{I}$ and $\bm E_{att}$:
\begin{equation}\small
	\begin{aligned}
		\ell_{cov\_id}=-\frac{2}{n_{b}}(\sum_{i=1}^{n_{b}/2}\bm q(\log(\bm W_{def}(\bm f_{i}^{V}))+\log(\bm W_{def}(\bm f_{i}^{I})))
	\end{aligned},
\end{equation}
where $\bm W_{def}$ is a pre-trained person identity classifier, $\bm q=(\frac{1}{M},\frac{1}{M}, \ldots ,\frac{1}{M})^T$, $M$ is the total number of person identifies in the training set, $n_b$ is the number of video sequences in a batch, and
\begin{equation}\small
	\begin{aligned}
		\bm f_{i}^{l}= \textrm{BN}(\textrm{GAP}(\bm E_{att}(\hat{\bm F}_{1, i}^{l}, \hat{\bm F}_{2,i}^l, \cdots, \hat{\bm F}_{T,i}^l)))
	\end{aligned},
\end{equation}
where $\hat{\bm F}_{t,i}^l(t = 1, 2 \cdots, T; l = V, I; i = 1, 2, \cdots, {n_b}/2)$ is the feature map of the $t$-th frame of the $i$-th sequence in the modality $l$ output by IMSA.

Minimizing Eq. (3) would activate the perturbations in the person image. In this paper, the perturbations are regarded as adversarial attack information. The activation helps improve the disturbing immunity of the defense network. In order to make the re-ID model robust to the diversity of pedestrian appearance features, $\bm f_{i}^{l}$ has been expected to practice adversarial attack and also related to the person identity. $\bm E^{V}$, $\bm E^{I}$, $\bm E_{att}$ are further updated by:
\begin{equation}\small
	\begin{aligned}
		\ell_{att\_id}= - \frac{2}{n_b}\bigg(\sum_{i = 1}^{n_b/2} \bm q_i (\log (\bm W_{att}(\bm f_{i}^{V}))+ \log (\bm W_{att}(\bm f_{i}^{I})))\bigg)
	\end{aligned},
\end{equation}
where $\bm q_i$ is a one-hot vector representing the identity of $\bm f_{i}^{V}$ and $\bm f_{i}^{I}$. $\bm W_{att}$ is the person identity classifier only used in ASAM.

\subsection{Adversarial Defense Module}
\begin{figure}[t!]
	\centering
	\includegraphics[width=1.0\linewidth]{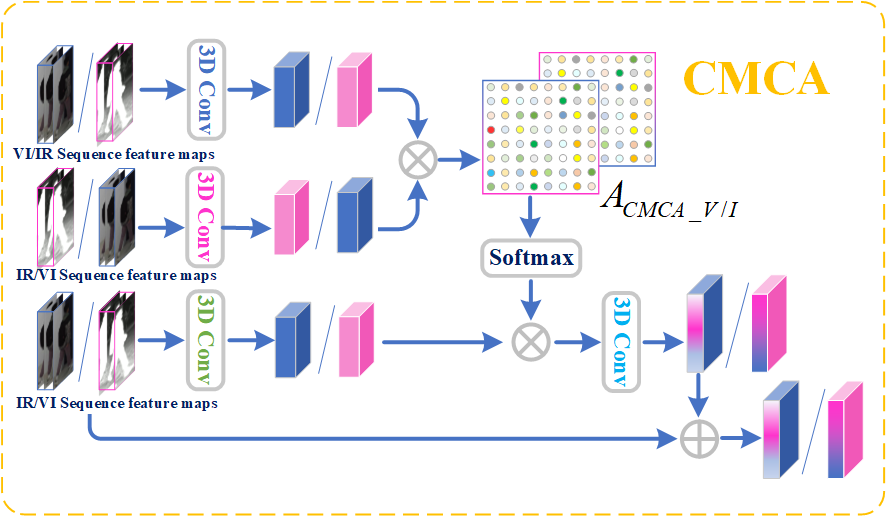}
	\caption{Structure of Cross-Modality Cross-Attention (CMCA) layer. $\otimes$ and $\oplus$ represent matrix multiplication and addition, respectively.}
	\label{fig4}
\end{figure}

ASAM is to activate the perturbations in the training samples and replace the role of the adversarial samples in the adversarial training to improve the robustness of the defense network $\bm E_{def}$ against the perturbations. To make $\bm E_{def}$ more immune to attack from the perturbations, an Adversarial Defense Module (ADM) is designed. ADM is mainly composed of defense network $\bm E_{def}$, cross-modality cross-attention (CMCA) layer, GAP and BN layers, as shown in Fig.~\ref{fig2}. The main task of the ADM is to
endow $\bm E_{def}$ with strong defense ability against the perturbations. In cross-modality person re-ID, modal-invariant features play a positive role in promoting the matching accuracy of person identities. Therefore, the features extracted by $\bm E_{def}$ contain rich information on different modalities, which would be helpful to defend against attacks from feature perturbation. A CMCA layer is embedded in the ADM, as shown in Fig.~\ref{fig4}.

As shown in Fig.~\ref{fig2}, there are two CMCA layers in the ADM, one embedded after the third convolution layer of $\bm E_{def}$, and the other embedded after the last convolution layer of $\bm E_{def}$. $\bm E_{def3}$ is composed of the first three convolution layers of $\bm E_{def}$ as an encoder. After the perturbations is activated by the ASAM, the feature maps $\bm F_{t,c}^V$ and $\bm F_{t,c}^I$ of the $t$-th frame are sent to the encoders $\bm E_{def3}$ and $\bm E_{def}$, the results are:
\begin{equation}\small
	\begin{aligned}
		\bm F_{t,d3}^V=\bm E_{def3}(\bm F_{t,c}^V), \bm F_{t,d3}^I=\bm E_{def3}(\bm F_{t,c}^I)
		\\\bm F_{t,d}^V=\bm E_{def}(\bm F_{t,c}^V),
		\bm F_{t,d}^I=\bm E_{def}(\bm F_{t,c}^I)
	\end{aligned}.
\end{equation}
After  $\bm F_{t, d3}^V$, $\bm F_{t, d3}^I$, $\bm F_{t, d}^V$ and $\bm F_{t, d}^I$ are input into CMCA layer, the results can be expressed as:
\begin{equation}\small
	\begin{aligned}
		&\bar{\bm F}_{t,d3}^V = \textrm{ConLa}_{4} (\textrm{CMCA}(\bm F_{t, d3}^V, \bm F_{t, d3}^I))
		\\&\bar{\bm F}_{t,d3}^I = \textrm{ConLa}_{4} (\textrm{CMCA}(\bm F_{t, d3}^I, \bm F_{t, d3}^V))
		\\&\bar{\bm F}_{t,d}^V = \textrm{CMCA}(\bm F_{t, d}^V, \bm F_{t, d}^I)
		\\&\bar{\bm F}_{t,d}^I = \textrm{CMCA} (\bm F_{t, d}^I, \bm F_{t, d}^V)
	\end{aligned},
\end{equation}
where $\rm{ConLa_{4}}$ denotes the last convolution layer of ${\bm {E}_{{def}}}$.

The common information can be extracted by embedding the CMCA layer in $\bm E_{def}$. The first CMCA layer enables ${\bm {E}_{{def}}}$ to extract discrimination feature maps with the common information on a shallow convolution layer. The second CMCA layer is used to ensure that the feature maps extracted by ${\bm {E}_{{def}}}$ contain common information for identity matching. To integrate the complementary information existing in ${\bar{\bm {F}}}_{t,{d}3}^l$ and ${\bar{\bm{F}}}_{t,{d}}^l$ ($l=V,I$) and realize the accurate description of person appearance features, we fuse ${\bar{\bm {F}}}_{t,{d}3}^l$ and ${\bar{\bm{F}}}_{t,{d}}^l$ $(l=V,I)$, respectively, and the fused results are sent to the GAP and BN layers.
The feature vectors obtained are:
\begin{equation}\small
	\begin{aligned}
		&{\bm{f}}_{d}^V = \textrm{BN}(\textrm{GAP}(({\bar{\bm{F}}}_{1,{d}3}^V + {\bar{\bm{F}}}_{1,{d}}^V)/2, \cdots, ({\bar{\bm{ F}}}_{T,{d}3}^V + {\bar{\bm{F}}}_{T,{d}}^V)/2))
		\\&{\bm{f}}_{d}^I = \textrm{BN}(\textrm{GAP}(({\bar{\bm{F}}}_{1,{d}3}^I + {\bar{\bm{F}}}_{1,{d}}^I)/2, \cdots ,({\bar{\bm{F}}}_{T,{d}3}^I + {\bar{\bm{F}}}_{T,{d}}^I)/2))
	\end{aligned}.
\end{equation}

To ensure that ${\bm{f}}_{d}^V$ and ${\bm{f}}_{d}^I$ have strong discrimination, we use the identity loss to optimize ${\bm{E}}_{{def}}$:
\begin{equation}\small
	\begin{aligned}
		\ell_{def\_id} =- \frac{2}{n_b}\bigg(\sum\limits_{i = 1}^{{n_b}/2} {\bm q_i}(\log (\bm{W}_{def}({\bm{f}}_{d,i}^V))+ \log (\bm W_{def}(\bm f_{d,i}^I))
		)\bigg)
	\end{aligned},
\end{equation}
where ${\bm{f}}_{{d, i}}^V$ and ${\bm{f}}_{{d,i}}^I$ represent the features of the $i$-th sequence in the visible and infrared modality, respectively. The triplet loss is used to solve the hard samples problem:
\begin{equation}\small
	\begin{aligned}
		{\ell_{def\_tri}}= \frac{1}{{{n_b}}}{\sum\limits_{i = 1}^{{n_b}} {\left[ {\left\| {{\bm{f}}_{{d,i}}^a - {\bm{f}}_{{d,i}}^p} \right\|_2^2 - \left\| {{\bm{f}}_{{d,i}}^a - {\bm{f}}_{{d,i}}^n} \right\|_2^2 + \alpha } \right]} _ + }
	\end{aligned},
\end{equation}
where $[\bullet]_{+}=\max\{0, \bullet\}$, ${\bm{f}}_{{d,i}}^a$ and ${\bm{f}}_{{d,i}}^p$ represent the features of the $i$-th anchor sample  sequence and the hard positive sample  sequence with the same identity in a mini-batch. ${\bm{f}}_{{d,i}}^n$ is a hard negative sample sequence with different identity from ${\bm{f}}_{d,i}^{a}$. $\alpha$ denotes the margin ($>0$, empirically set to 0.3 in this work). The features ${\bm{f}}_{{d,i}}^a$, ${\bm{f}}_{{d,i}}^p$  and ${\bm{f}}_{{d,i}}^n$ are generated by Eq. (8).

\subsection{Feature Representation under Spatial-temporal Information Guidance}
\begin{figure*}[htbp]
	\centering
	\includegraphics[width=1.0\textwidth]{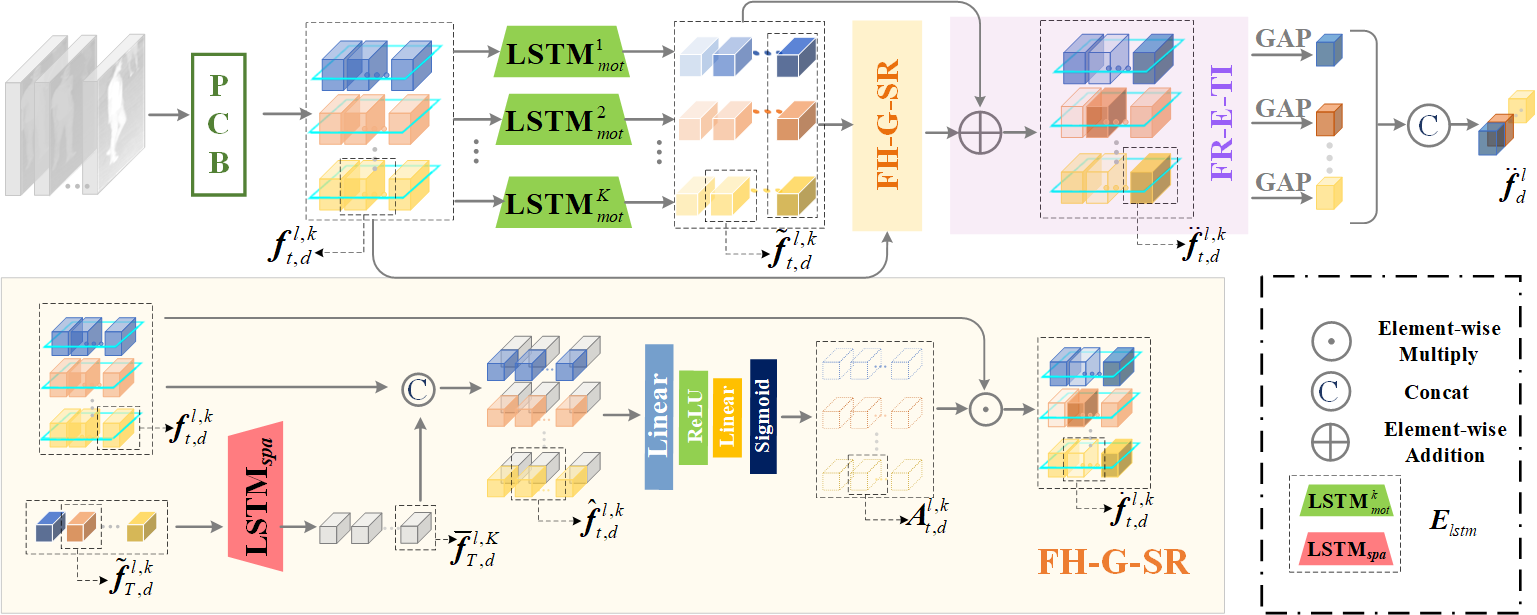}
	\caption{Feature Representation under Spatial-temporal Information Guidance.}
	\label{fig5}
\end{figure*}

The video sequence of a person contains a lot of motion information, which is not affected by the modality changes. In a single pedestrian image, there is a latent spatial relation between different regions of the pedestrian's body, and different pedestrians usually show different relations. In order to guide the discrimination features learning, a spatial-temporal information mining approach is proposed, as shown in Fig.~\ref{fig5}.

The proposed method is mainly composed of two parts: 1) feature highlighting guided by spatial relation (FH-G-SR) and 2) feature representation embedded by temporal information (FR-E-TI). To highlight the discrimination feature with spatial relation guidance, we first use PCB~\cite{61} to divide the feature map $\bm F_{t, d}^V$ and $\bm F_{t, d}^I$ into $K$ different patches in space, which are converted into feature vectors. The feature vectors of the $k$-th patches of ${\bm F}_{t,{d}}^V$ and ${\bm F}_{t,{d}}^I$ are denoted as ${\bm f}_{t, d}^{V, k}$ and ${\bm f}_{t, d}^{I, k}$ ($k = 1,2, \cdots, K$). Based on the experience of PCB, $K$ is set to 6 in this paper. As can be seen in Fig.~\ref{fig5}, the video sequence features $\{ {\bm f}_{1, d}^{l, k}, \cdots, {\bm f}_{T, d}^{l,k}\}$ are sent into $\textrm{LSTM}_{mot}$ to obtain the features embedded with motion information, expressed as $\{\tilde{\bm f}_{1, d}^{l,k}, \cdots , \tilde{\bm f}_{T, d}^{l,k}\}$. Since the potential spatial relation between patches at the different positions is not involved, a sequence of different patches (of the same frame) is formed and input into $\textrm{LSTM}_{spa}$ for spatial relationship mining between different patches. Considering that after the last frame passes through $\textrm{LSTM}_{mot}$, the obtained features integrate the information of all previous frames, we only form a sequence $\{\tilde{\bm f}_{T, d}^{l,1}, \cdots , \tilde{\bm f}_{T, d}^{l,K}\}(l = V, I)$ of all patch features of the $T$-th frame and send it to $\textrm{LSTM}_{spa}$.

\subsubsection{FH-G-SR}
The result $\bar{\bm f}_{T, d}^{l, K}$ obtained after feeding $\{\tilde{\bm f}_{T, d}^{l,1}, \cdots , \tilde{\bm f}_{T, d}^{l,K}\}(l = V, I)$ into $\textrm{LSTM}_{spa}$ is the final spatial feature representation. To effectively utilize the spatial relations and the information carried by $\bm f_{t,d}^{l,k}$ to highlight discrimination features, we concatenate $\bar{\bm f}_{T, d}^{l, K}$ and the original visual feature $\bm f_{t, d}^{l, k}$:
\begin{equation}\small
	\begin{aligned}
		\hat{\bm f}_{t,d}^{l,k}=\textrm{concat}(\bm f_{t,d}^{l,k}, \bar{\bm f}_{T,d}^{l,k})
	\end{aligned}.
\end{equation}
As shown in Fig.~\ref{fig5}, after $\hat{\bm f}_{t,d}^{l,k}$ passes through the linear mapping (LM) layer,  ReLU activation function, the other LM layer, and the Sigmoid activation function, the corresponding weight matrix for features highlighting is obtained by:
\begin{equation}\small
	\begin{aligned}
		\bm A_{t, d}^{l, k}=\textrm{Sigmoid}(\textrm{LM}(\textrm{ReLU}(\textrm{LM}(\hat{\bm f}_{t,d}^{l,k}))))
	\end{aligned}.
\end{equation}
With $ \bm A_{t, d}^{l, k}$, the feature $\bm f_{t, d}^{l, k}$ highlighted by spatial relation guidance is:
\begin{equation}\small
	\begin{aligned}
		\dot{\bm f}_{t, d}^{l, k}=\bm f_{t, d}^{l, k}\odot\bm A_{t, d}^{l, k}
	\end{aligned}.
\end{equation}

\subsubsection{FR-E-TI}
Although $\dot{\bm f}_{t, d}^{l, k}$ makes use of the spatial relation between patches, it is only the visual feature of a person, and it does not integrate the motion information contained in the video sequence. Therefore, we embedded the features $\{\tilde{\bm f}_{1, d}^{l,k}, \cdots, \tilde{\bm f}_{T, d}^{l,k}\}$ carrying motion information into the enhanced features in Eq. (13):
\begin{equation}\small
	\begin{aligned}
		\ddot{\bm f}_{t, d}^{l,k} = \dot{\bm f}_{t, d}^{l, k} + {\tilde{\bm f}}_{t, d}^{l,k}
	\end{aligned}.
\end{equation}
GAP is used to achieve the fusion of $T$ frame features $\{\ddot {\bm f}_{1, d}^{l, k}, \cdots, \ddot{\bm f}_{T, d}^{l, k}\}$, and the feature representation of the $k$-th patch of modality $l$ can be obtained:
\begin{equation}\small
	\begin{aligned}
		\ddot{\bm f}_{d}^{l,k} = \textrm{GAP}(\ddot{\bm f}_{1, d}^{l,k}, \cdots, \ddot{\bm f}_{T, d}^{l,k})
	\end{aligned}.
\end{equation}

Finally, we concatenate the features of all patches together according to their spatial positions on the image to form a complete person representation $\ddot{\bm f}_{d}^l ~(l = V, I)$. In order to ensure its discrimination, the cross entropy loss is deployed:
\begin{equation}\small
	\begin{aligned}
		{\ell_{p\_id}}= - \frac{2}{n_b}\sum_{i = 1}^{n_b/2} {\bm q_i}\log \left( {\bm W_{se}\big( {{\ddot{\bm f}}_{{d, i}}^{V}} \big)} \right)+{\bm q_i}\log \left( {\bm W_{se}\big( {{\ddot{\bm f}}_{{d, i}}^{I}} \big)} \right)
	\end{aligned},
\end{equation}
where $\bm W_{se}$ is the identity classifier of $\ddot{\bm f}_{d, i}^{l}$ which is generated via Eq. (15). $\ddot{\bm f}_{d, i}^{l}$ denotes the sequence feature of the $i$-th video sequence in one batch.

The overall loss function in the proposed approach is:
\begin{equation}\small
	\begin{aligned}
		\ell_{total} &= \ell_{cov\_id}+ \lambda _1 \ell_{att\_id}+  \lambda _2 (\ell_{def\_id}+  \ell_{def\_tri}) +\lambda _3 \ell_{p\_id}
	\end{aligned},
\end{equation}
where ${\lambda _1}$, ${\lambda _2}$, ${\lambda _3}$ are hyper-parameters, which are used to adjust the role of the corresponding loss items. The processes are summarized in Algorithm~\ref{alg:A1}.

\begin{algorithm}[htbp]
	\caption{\textbf{} Self-Attack Defense and Spatial-Temporal Relation Mining for Visible-Infrared Video Person Re-ID}\label{alg:A1}
	\begin{algorithmic}\small
				\State {\textbf{Input:} Visible image sequence $\bm V = \{ {V_t}\} _{t = 1}^T$, infrared image sequence $\bm I = \{ {I_t}\} _{t = 1}^T$. Maximum number of iterations \emph{iter1 } and \textit{iter2}.
				\State {\textbf{Output:} The trained  $\bm E_{def}$, $\bm E^{V}$ and $\bm E^{I}$.\\
				\begin{flushleft}
					\textbf{Step \uppercase\expandafter{\romannumeral1}:} Training of ASAM. \\
					~1:\textbf{for} $iter=1$, $\cdots$, $iter1$  \textbf{do}\\
					~2:\qquad Update $\bm E^{V}$, $\bm E^{I}$, $\bm E_{def}$, $\bm W_{def}$ by minimizing Eqs. (9)\\~~~~\qquad and  (10).\\
					~3:\qquad Update $\bm E^{V}$, $\bm E^{I}$, $\bm E_{def}$, $\bm E_{lstm}$ and $\bm W_{se}$ by minimizing\\~~~~\qquad Eq. (16).\\
					~4:\qquad Fixed the parameters of $\bm W_{def}$.\\
					~5:\qquad Update $\bm E_{def}$, $\bm E^{V}$, $\bm E^{I}$, $\bm E_{att}$ and $\bm W_{att}$ by minimizing \\~~~~\qquad Eqs. (3)  and (5).\\
					~6:\textbf{end for}\\
					\textbf{Step \uppercase\expandafter{\romannumeral2}:} Training of ADM and FRM-STIG.\\
					~7:\textbf{for} $iter=1$, $\cdots$, $iter2$  \textbf{do}\\
					~8:\qquad Fixed the parameters of $\bm W_{def}$.\\
					~9:\qquad Update $\bm E_{def}$ by minimizing Eqs. (9) and (10).\\
					10:\qquad Update $\bm E^{V}$, $\bm E^{I}$, $\bm E_{def}$, $\bm E_{lstm}$ and $\bm W_{se}$ by minimizing\\~~~~\qquad Eq. (16).\\
					11:\qquad Update $\bm E^{V}$, $\bm E^{I}$, $\bm E_{att}$ and $\bm W_{att}$ by minimizing Eqs. (3),\\~~~~\qquad (5).\\
					12:\textbf{end for}\\
		\end{flushleft}}}
	\end{algorithmic}
\end{algorithm}

\section{Experiments}\label{sec4}
\subsection{Experimental Settings}
The dataset used in this experiment is a large-scale cross-modal video-based person re-ID dataset---VCM proposed by Lin et al.~\cite{30}, which is the first and only one currently constructed for the visible-infrared video person re-ID task. The dataset is recorded by 12 non-overlapping HD cameras and consists of 251,452 visible images and 211,807 infrared images with a resolution of $3,840 \times 2,170$. These images are further divided into 11,785 sequences in the visible modality and 10,078 sequences in the infrared modality. The dataset contains 927 identities, where 232,496 images of 500 identities involving a total of 11,061 sequences are used for training, and the remaining 230,763 images of 427 identities involving a total of 10,802 sequences are used for testing.

All experiments are carried out on a PC equipped with an NVIDIA TESLA A100 GPU in the Pytorch 1.10 framework~\cite{52}. In the training phase, all input images are adjusted to $288 \times 144$. The batch size is set to 32 (i.e., 32 sequences are processed in a mini-batch). In each epoch, 16 sequences of each modality enter the model for training (containing eight identities, each containing two sequences). Each sequence consists of 6 frames, a total of 192 frames. The model is trained for 200 epochs (each of which contains 268 iterations). The first 150 epochs are used to train $\bm E^{V}$, $\bm E^{I}$, $\bm E_{att}$, $\bm E_{lstm}$, $\bm W_{att}$ and $\bm W_{def}$. For the remaining 50 epochs, we fix $\bm W_{def}$ and fine-tune $\bm E_{def}$ to further enhance the network's defense capability. The entire training is realized by using SGD optimizer with the momentum of 0.9, weight decay of $5 \times 10^{-4}$ and learning rate of 0.12. A warm-up strategy~\cite{74} is applied to tune the learning rate linearly. Cumulative Matching Curve (CMC)~\cite{53} and mean Average Precision (mAP)~\cite{54} are used as the evaluation metrics for model performance.

\begin{table*}[htbp]\footnotesize
	\centering {\caption{The results of ablation experiment. *denotes the model without LSTMspa} \label{table1}
		\renewcommand\arraystretch{1.3}
		\begin{tabular}{ccccccccccc}
			\toprule
			\multirow{2}*{Methods} & \multicolumn{5}{c}{Infrared to Visible} & \multicolumn{5}{c}{Visible to Infrared} \\
			\cline{2-11}
			&Rank-1&Rank-5 &Rank-10&Rank-20&mAP&Rank-1&Rank-5 &Rank-10&Rank-20&mAP\\
			\hline
			Baseline    &58.05  &72.45  &78.49  &83.90  &43.23  &61.01  &74.98   &80.41  &83.92  &45.59\\
			Baseline+ASAM   &43.65  &62.37  &70.38	 &77.55  &30.66   &46.01  &66.58   &74.17	 &81.39  &30.63\\
			Baseline+ADM   &58.92  &73.01  &78.42	 &83.77  &44.50   &62.58  &77.03   &81.88	 &86.27  &46.55\\
			Baseline+FRM-STIG      &61.13  &76.00  &81.17  &85.89  &44.80  &63.74  &78.05   &83.02  &87.43  &46.34\\
			Baseline+ASAM+ADM    &60.27  &74.78  &80.63  &85.51  &46.09  &63.01  &77.19   &82.21  &86.41  &48.05\\
			Baseline+ASAM+ADM+FR-E-TI  &63.92 &77.47 &82.61 &87.04 &48.56 &66.42 &79.92 &84.53 &88.78 &50.40\\
			Baseline+ASAM+ADM+FRM-STIG*    &64.07  &77.44  &83.25  &87.35  &48.79  &65.78  &79.60  &84.60  &88.66  &50.57\\
			Baseline+ASAM+ADM+FRM-STIG     &\bf65.31  &\bf77.86  &\bf82.66  &\bf87.35  &\bf49.49  &\bf67.66  &\bf80.74   &\bf85.13  &\bf89.14  &\bf51.76\\
			\botrule
	\end{tabular}}
\end{table*}

\begin{figure*}[htbp]
	\centering
	\includegraphics[width=1.0\textwidth]{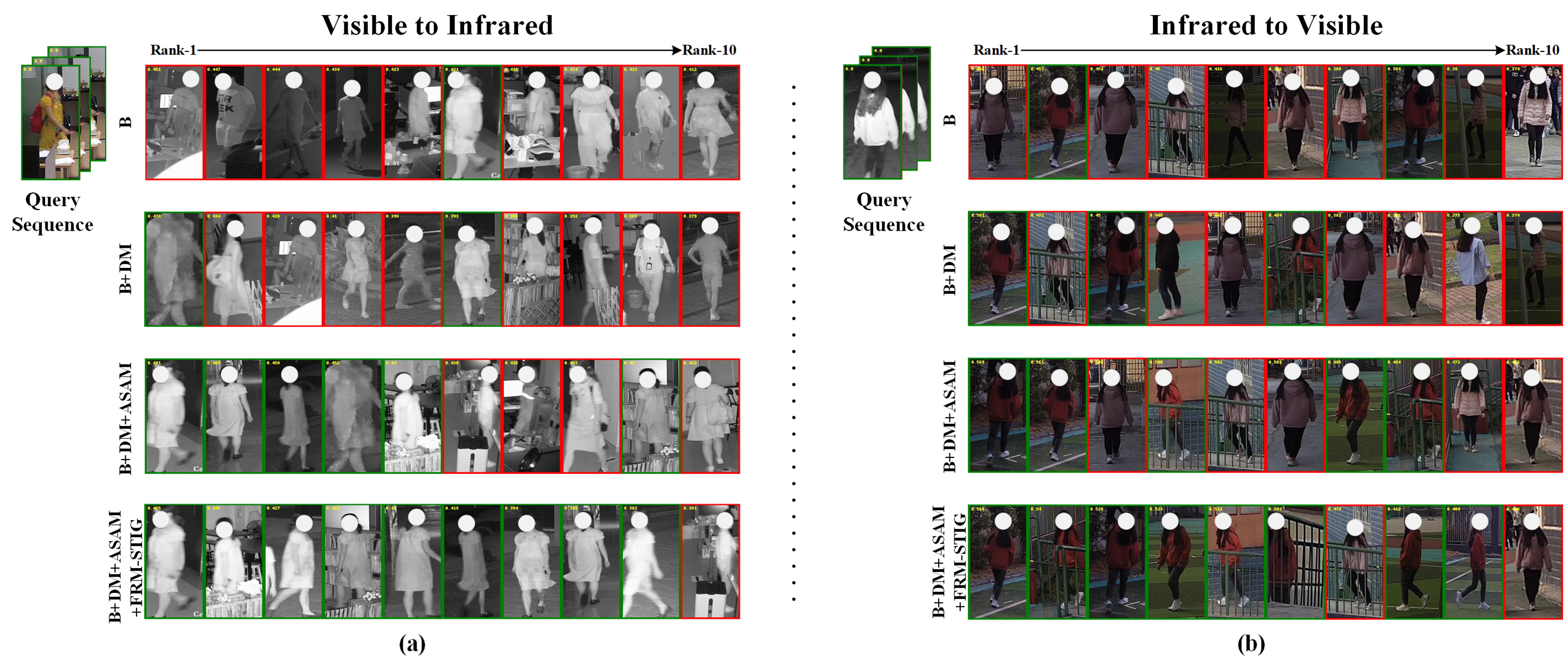}
	\caption{Visualization of retrieval results of Rank-1$\rightarrow$Rank10 under different experimental settings. ``B'' denotes the baseline. (a) Results on the task of ``Visible to Infrared''. (b) Results on the task of ``Infrared to Visible''.}
	\label{fig6}
\end{figure*}

\subsection{Ablation Study}
The proposed method consists of an adversarial self-attack module (ASAM), adversarial defense module (ADM), and feature representation module under spatial-temporal information guidance (FRM-STIG). $\bm E^{V}$, $\bm E^{I}$ and $\bm E_{def}$ trained by cross-entropy loss and triplet loss are regarded as ``Baseline''. The ``Baseline'' is pre-trained on the ImageNet~\cite{55} before training on the dataset VCM. We denote the method of adding ASAM to the baseline as ``Baseline+ASAM'', similarly, we obtain ``Baseline+ADM'', ``Baseline+FRM-STIG'' and ``Baseline+ASAM+ADM''. When FH-G-SR is removed from FRM-STIG, and the remaining content is added to ``Baseline+ASAM+ADM''. Such method is noted ``Baseline+ASAM+ADM+FR-E-TI''. The complete proposed model is marked ``Baseline+ASAM+ADM +FRM-STIG''. Furthermore, in order to verify the contribution of $\textrm{LSTM}_{spa}$, $\textrm{LSTM}_{spa}$ is removed from ``Baseline+ASAM+ADM+FRM-STIG'' and the corresponding model is denoted as ``Baseline+ASAM+ADM+FRM-STIG*''. The results of ablation experiment are reported in Table~\ref{table1}.

\textbf{Effectiveness of ASAM}. To verify the effect of the ASAM, we add the ASAM to the ``Baseline'', and obtain the model  ``Baseline+ASAM''. One can see in Table~\ref{table1} that, on the ``Infrared to Visible'' task, Rank-1 and mAP achieved by ``Baseline+ASAM'' decrease by 14.4\% and 12.57\%, respectively. For the task of ``Visible to Infrared'', Rank-1 and mAP are reduced by 15\% and 14.96\%, respectively. These indicate the attacks from the perturbations activated in the shallow features.

\textbf{Effectiveness of ADM}. In Table~\ref{table1}, Rank-1 and mAP accuracy of ``Baseline+ADM'' on the task of querying the visible sequence from the infrared sequence (i.e., Infrared to Visible) is 58.92\% and 44.50\%, respectively. Compared with that of ``Baseline'', the performance is improved by 0.87\% and 1.27\%, respectively. On the task of ``Visible to Infrared'' task, the accuracy of Rank-1 and mAP reaches 62.58\% and 46.55\%, respectively. Compared with that of  ``Baseline'', the performance of ``Baseline+ADM'' is improved by 1.57\% and 0.96\%. It implies that the ADM still has a positive effect on the model performance improvement when the ASAM is absent. It can be also observed that when ASAM and ADM are added to ``Baseline'' together, the performance of ``Baseline+ASAM+ADM'' is improved. Compared with ``Baseline+ADM'', Rank-1 and mAP of ``Baseline+ASAM+ADM'' on ``Infrared to Visible'' task are increased from 58.92\% and 44.50\% to 60.27\% and 46.09\%. On ``Visible to Infrared'' task, the performance of Rank-1 and mAP are improved from 62.58\% and 46.55\% to 63.01\% and 48.05\%. These demonstrates the effectiveness of the adversarial training.

\textbf{Effectiveness of FR-E-TI}. FH-G-SR is removed from FRM-STIG to evaluate the validity of FR-E-TI with temporal information embedding. It can be seen in Table~\ref{table1} that when the FR-E-TI is added to ``Baseline+ASAM+ADM'', Rank-1 and mAP of the model ``Baseline+ASAM+ADM+FR-E-TI'' on ``Infrared to Visible'' (``Visible to Infrared'') task are improved from 60.27\% and 46.09\% (63.01\% and 48.05\%) to 63.92\% and 48.56\% (66.42\% and 50.40\%), respectively. The improvement verifies the validity of FR-E-TI when FH-G-SR is absent.

\textbf{Effectiveness of FRM-STIG}.
It can be seen in Table~\ref{table1} that with FRM-STIG, the performance of ``Baseline+FRM-STIG''  on the ``Infrared to Visible'' (``Visible to Infrared'') task, the accuracy of Rank-1 and mAP increases from 58.05\% and 43.23\% (61.13\% and 44.80\%) to 61.01\% and 45.59\% (63.74\% and 46.34\%) respectively. For the same task, after FRM-STIG is added to ``Baseline+ASAM+ADM'', the accuracy of the model ``Baseline+ASAM+ADM+FRM-STIG'' on Rank-1 and mAP increases from 60.27\% and 46.09\% (63.01\% and 48.05\%) to 65.31\% and 49.49\% (67.66\% and 51.76\%) respectively. It verifies the contribution of FRM-STIG. Moreover, compared with the performance of ``Baseline+ASAM+ADM+FR-E-TI'',  Rank-1 and mAP of ``Baseline+ASAM+ADM+FRM-STIG'' on ``Visible to Infrared'' (``Infrared to Visible'') are improved from 63.92\%  and 48.56\% (66.42\% and 50.40\%) to 65.31\% and 49.49\% (67.66\% and 51.76\%). It demonstrates the validity of FH-G-SR. Further, Rank-1 and mAP of ``B+ASAM+ADM+FRM-STIG*'' on ``Visible to Infrared'' (``Infrared to Visible'')  decrease by 1.24\% and 0.7\% (1.88\% and 1.19\%) compared with those of ``Baseline+ASAM+ADM+FRM-STIG''. It verifies the contribution of LSTM$_{spa}$.

\begin{figure}[htbp]
	\centering
	\includegraphics[width=1.0\linewidth]{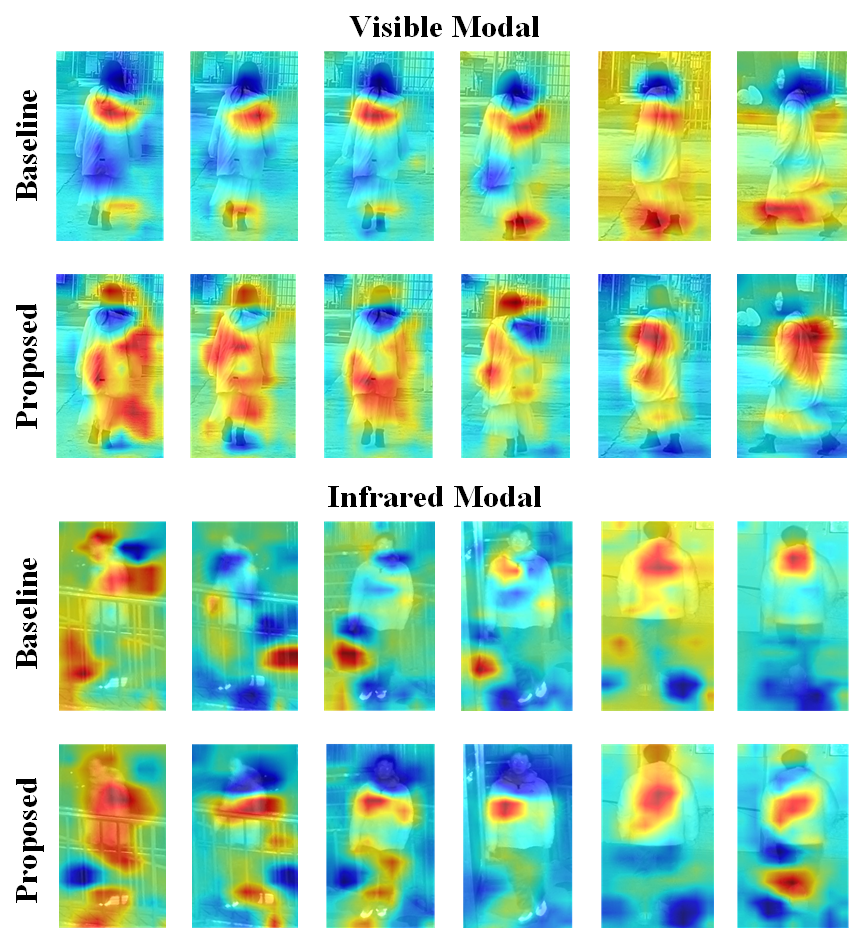}
	\caption{Comparison of the areas focused by Baseline and the proposed method.}
	\label{fig7}
\end{figure}

The visual effect of different settings in the ablation experiment on the retrieval results is shown in Fig.~\ref{fig6}. From the results shown in Fig.~\ref{fig6}, one can
see that the retrieval accuracy has improved when the ADM is added to the ``Baseline''. It found that when ASAM and ADM are added to the ``Baseline'' together, the model
performance is visually improved. It indicates that the adversarial attack and defense strategies proposed are effective. Besides, when FRM-STIG is added to ``Baseline+ASAM+ADM'', the matching accuracy of sequences is further improved. Fig.~\ref{fig7} shows the areas focused by ``Baseline'' and the proposed method, where the warmer the color is, the more attention the area receives. Those results indicate that the proposed method can better extract discriminative features from the person's body area than the Baseline.

\begin{table*}[!ht]\footnotesize
	\centering {\caption{Performance comparison with state-of-the-art methods on VCM. The CMC and mAP evaluation results (\%) are listed. ``*'' denotes the FRM-STIG removed from the proposed complete model. The best result is shown in bold. }
		\label{table2}
		\renewcommand\arraystretch{1.3}
		\begin{tabular}{ccccccccccc}
			\toprule
			\multirow{2}*{Methods} & \multicolumn{5}{c}{Infrared to Visible} & \multicolumn{5}{c}{Visible to Infrared} \\
			\cline{2-11}
			&Rank-1&Rank-5 &Rank-10&Rank-20&mAP&Rank-1&Rank-5 &Rank-10&Rank-20&mAP\\
			\hline
			LbA\cite{56}(ICCV'21)    &46.38  &65.29  &72.23	 &79.41  &30.69     &49.30  &69.27  &75.90	 &82.21  &32.38\\
			MPANet\cite{17}(CVPR'21)    &46.51  &63.07  &70.51	 &77.77  &35.26      &50.32  &67.31  &73.56	 &79.66  &37.80\\
			DDAG\cite{58}(ECCV'20)    &54.62  &69.79  &76.05	 &81.50  &39.26     &59.03  &74.64  &79.53	 &84.04  &41.50\\
			VSD\cite{59}(CVPR'21)    &54.53  &70.01  &76.28	 &82.01  &41.18     &57.52  &73.66  &79.38	 &83.61  &43.35\\
			CAJL\cite{60}(ICCV'21)    &56.59  &73.49  &79.52	 &84.05  &41.49     &60.13  &74.62  &79.86	 &84.53  &42.81\\
			MITML~\cite{30}(CVPR'22)    &63.74  &76.88  &81.72	 &86.28  &45.31     &64.54  &78.96  &82.98	 &87.10  &47.69\\
			\hline
			\bf Proposed* &\bf60.27	&\bf74.78	&\bf80.63	&\bf85.51	&\bf46.09	&\bf63.01	&\bf77.19	&\bf82.21	&\bf86.41	&\bf48.05\\
			\bf Proposed &\bf65.31  &\bf77.86  &\bf82.66  &\bf87.35  &\bf49.49  &\bf67.66  &\bf80.74   &\bf85.13  &\bf89.14  &\bf51.76\\	
			\botrule
	\end{tabular}}
\end{table*}

\subsection{Comparison with State-of-the-Arts}
In order to verify the superiority of the proposed method over the existing methods, it is compared with LbA~\cite{56}, MPANet~\cite{17}, DDAG~\cite{58}, VSD~\cite{59}, CAJL~\cite{60}, MITML~\cite{30}, where the first five methods are designed for image-based visible-infrared person re-ID and the last one is for visible-infrared video person re-ID. Since the first five methods are proposed for the single-frame visible-infrared person image matching task, we remove FRM-STIG for comparison and such method ``Proposed*'' in Table~\ref{table2}. Rank1 and mAP of ``Proposed*'' are 60.27\% and 46.09\% (63.01\% and 48.05\%) on ``Infrared to Visible'' (``Visible to Infrared''), which are 3.68\% and 4.60\% (2.88\% and 5.24\%) higher than those of the sub-optimal image-based method CAJL. Compared with the latest video person re-ID method MITML, Rank-1 and mAP of the proposed method are increased from 63.74\% and 45.31\% (64.54\% and 47.69\%) to 65.31\% and 49.49\% (67.66\% and 51.76\%) on the ``Infrared to visible'' (``Visible to Infrared'') task. It shows that the proposed method outperforms all compared ones.

\begin{figure*}[htbp]
	\centering
	\subfigure []{\includegraphics[height=0.25\textwidth]{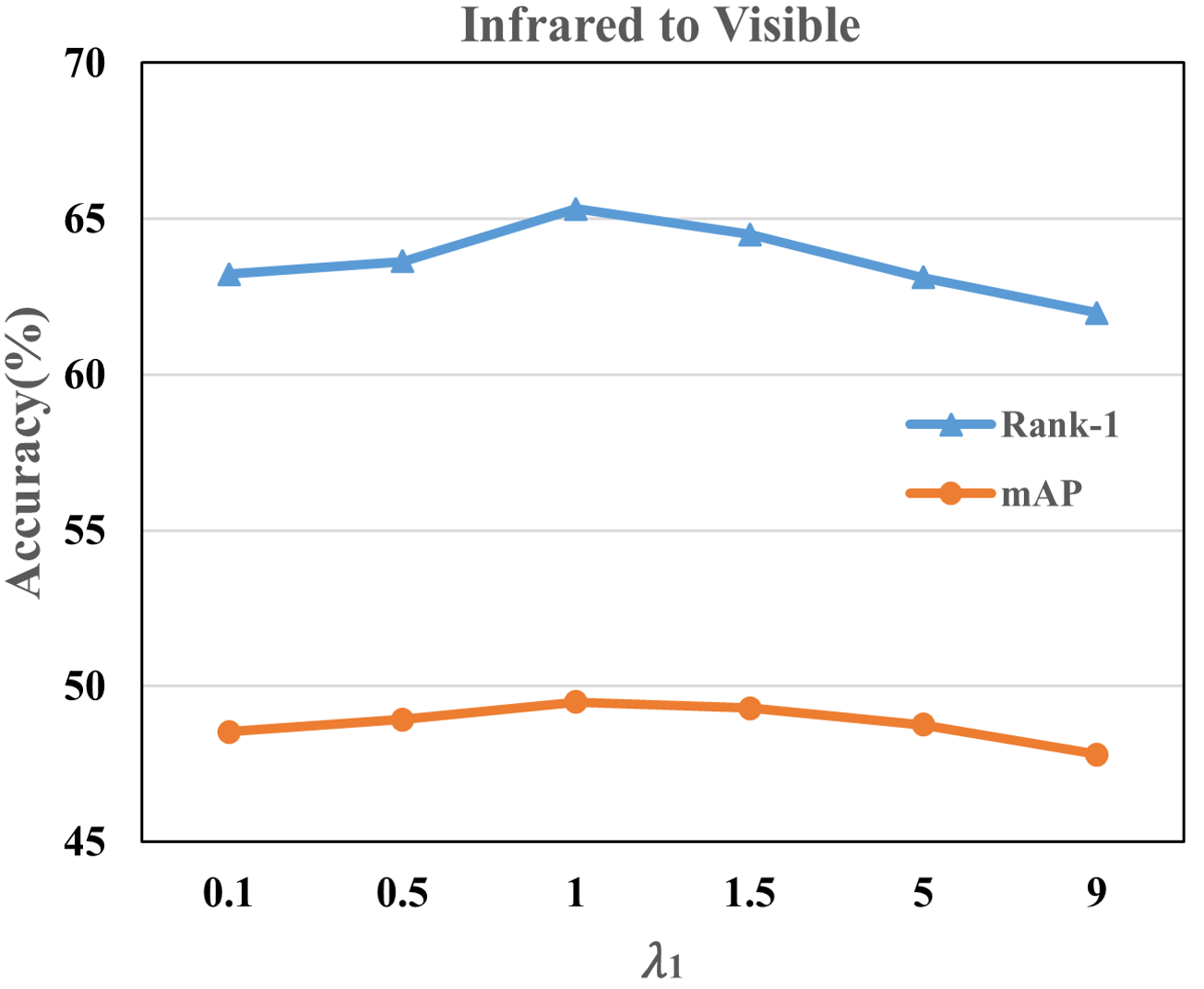}}
	\subfigure []{\includegraphics[height=0.25\textwidth]{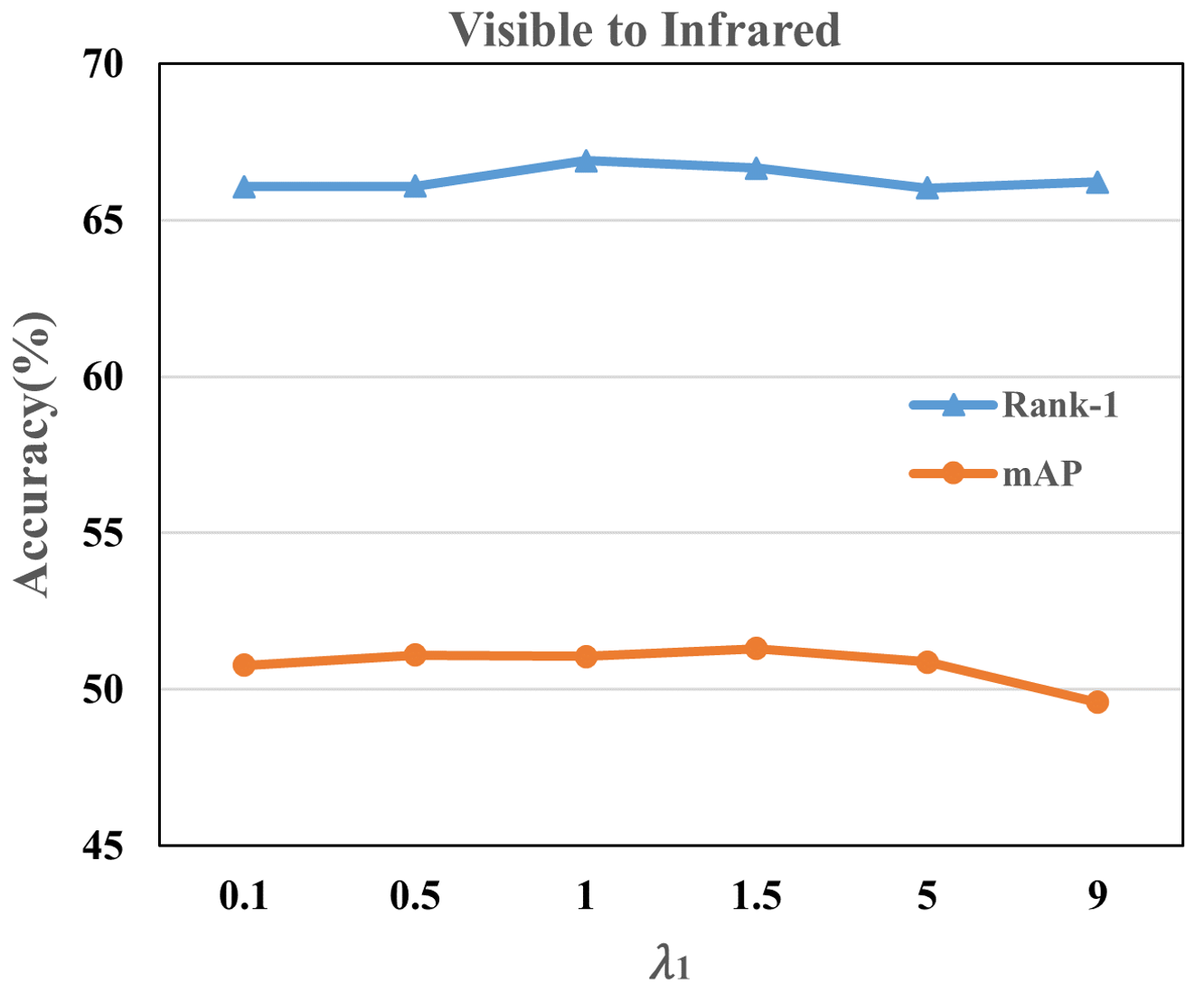}}
	\subfigure []{\includegraphics[height=0.25\textwidth]{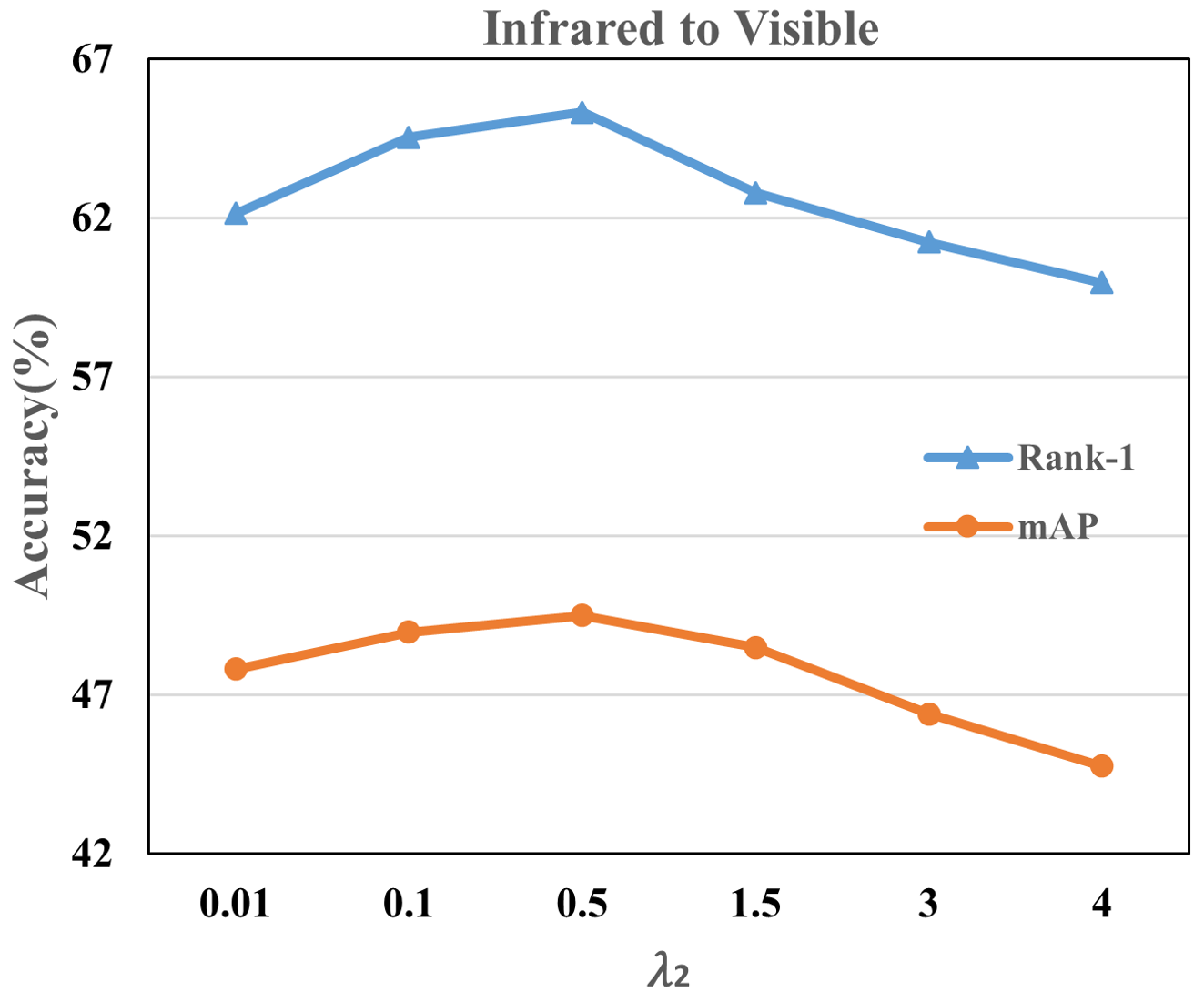}}
	\subfigure []{\includegraphics[height=0.25\textwidth]{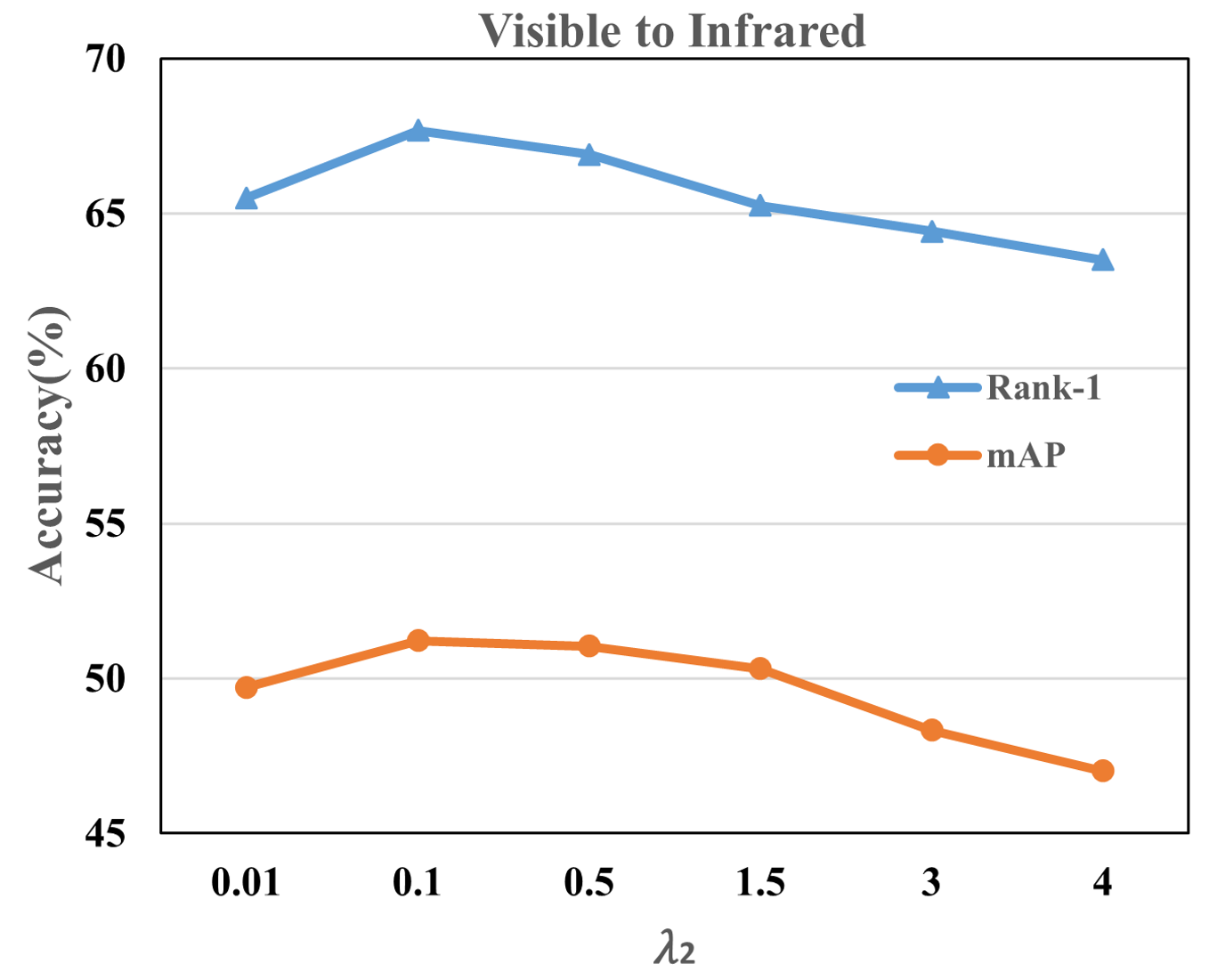}}
	\subfigure []{\includegraphics[height=0.25\textwidth]{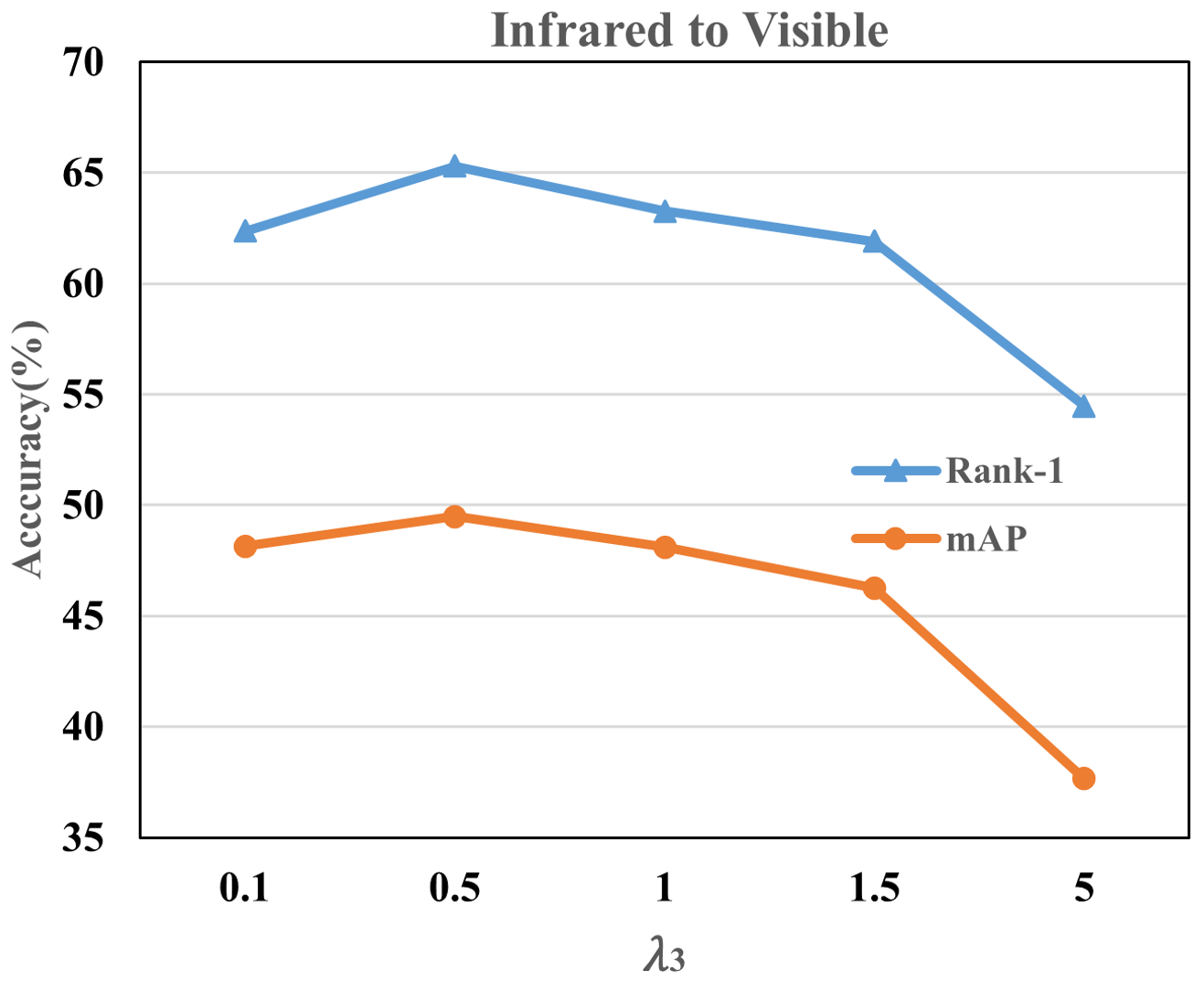}}
	\subfigure []{\includegraphics[height=0.25\textwidth]{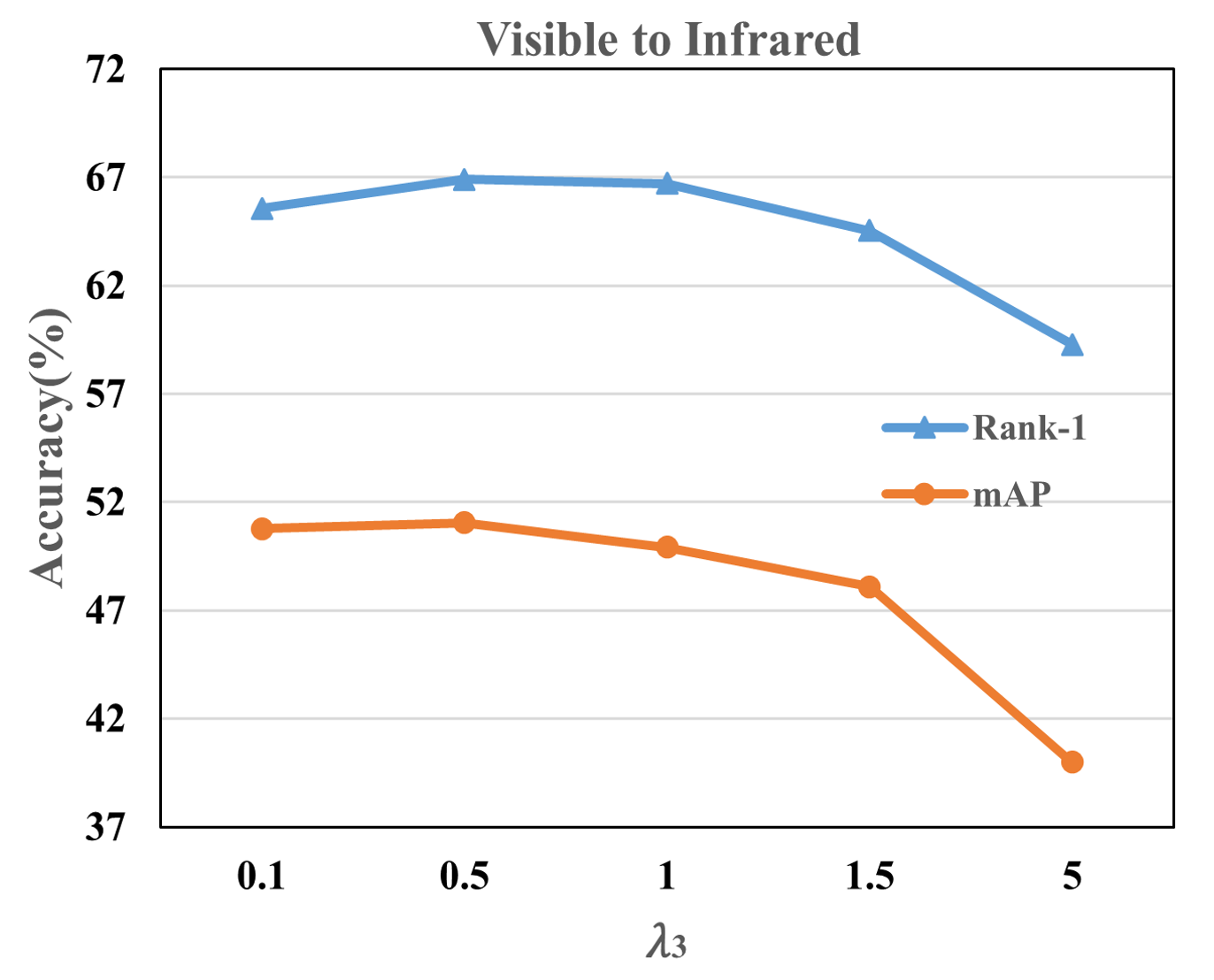}}
	\caption{Performance changes of proposed method with different $\lambda_{1}$, $\lambda_{2}$ and $\lambda_{3}$ values. (a) and (b) the influence of $\lambda_{1}$ when it changes from 0.1 to 9 on ``Infrared to Visible'' and ``Visible to Infrared'' task; (c) and (d) the influence of $\lambda_{2}$ when it changes from 0.01 to 4; (e) and (f) the influence of $\lambda_{3}$ when it changes from 0.1 to 5.}
	\label{fig8}
\end{figure*}

\subsection{Parameter Analysis}
In Eq. (17), three hyper-parameters $\lambda_{1}$, $\lambda_{2}$ and $\lambda_{3}$ need to be set. In this section, we discuss the influence of one hyper-parameter by fixing
the other two parameters. The performance of the proposed method with different hyper-parameters is shown in Fig.~\ref{fig8}.

\textbf{The influence of $\lambda_{1}$}.
Fig.~\ref{fig8} (a) and (b) show the effect of ${\lambda _1}$  when it changes in $[0.1,9]$. On the task of ``Infrared to Visible'', the proposed method achieves the best performance when ${\lambda _1} = 1$, and the performance degenerates when ${\lambda _1}>1$. On the task of ``Visible to Infrared'', the method in this paper shows insensitivity to the change of ${\lambda _1}$ value. Therefore, we set ${\lambda _1}$ to 1 in our method.

\textbf{The influence of $\lambda_{2}$}.
Fig.~\ref{fig8} (c) and (d) show the changes of the model performance when  ${\lambda _2}$ changes from 0.01 to 4. On task of ``Infrared to Visible'', when ${\lambda _2} = 0.5$ , the proposed method achieves the highest recognition accuracy. On ``Visible to Infrared'' task, when ${\lambda _2} = 0.1$, the proposed method achieves the highest recognition accuracy, and when ${\lambda _2} > 0.5$, the experimental performance performs a significant downward trend on both of  tasks. In this work, we set ${\lambda _2}$ to 0.5 for both tasks.

\textbf{The influence of $\lambda_{3}$}.
Fig.~\ref{fig8} (e) and (f) show the changes in performance of the proposed algorithm on  ``Infrared to Visible'' and ``Visible to Infrared'' tasks, when ${\lambda _3}$ varies between 0.1 and 5. As indicated in Fig.~\ref{fig8} (e) and (f), for both recognition tasks, the recognition performance of the proposed method reaches its peak when the value of $\lambda_{3}$ reaches 0.5. Therefore, we set $\lambda_{3}$ to 0.5 throughout the experiments.

\section{Conclusion}\label{sec5}
An adversarial self-attack defense and feature representation module under spatial-temporal information guidance is proposed for the diversity of pedestrian appearance features on pedestrian identity matching. The method consists of the ASAM, the ADM, and the FRM-STIG. Through the cooperative training of the ASAM and the ADM, the defense network's defense capability has been improved in the face of identity-related perturbation. The proposed method is robust to modality differences and feature changes caused by other factors. In addition, FRM-STIG utilizes each local feature effectively through a spatial relationship-guided highlight mechanism. The experimental results show that the proposed method outperforms the compared SOTA methods.

\bmhead{Acknowledgments}

This work was supported by the National Natural Science Foundation of China under Grants 62276120, 61966021 and 62161015.

\section*{Declarations}

\textbf{Conflict of interest} The authors declare that they have no known competing financial interests or personal relationships that could have appeared to influence the work reported in this paper.

\section*{Data Availability Statement}
The datasets for this study can be found in the VCM dataset \url{https://github.com/VCM-project233/MITML}.








\bibliography{mybibfile}


\end{document}